\PassOptionsToPackage{table}{xcolor}
\documentclass[sigconf]{acmart}

\usepackage{enumitem}
\usepackage{hyperref}
\usepackage{amsmath} 
\usepackage{multirow}
\usepackage{graphicx}
\usepackage{subfigure}
\usepackage{bm}
\usepackage{tabularx}
\usepackage{framed}
\usepackage{lipsum}
\usepackage{xcolor}
\usepackage{stfloats}

\usepackage{balance} 
\usepackage{enumitem}
\usepackage{hyperref}
\usepackage{amsmath} 
\usepackage{multirow}
\usepackage{graphicx}
\usepackage{subfigure}
\usepackage{bm}
\usepackage{tabularx}
\usepackage{framed}
\usepackage{xcolor}
\usepackage{stfloats} 
\usepackage{xspace}
\usepackage{comment}
\usepackage{multirow}
\usepackage{booktabs}
\usepackage{amsfonts}
\usepackage{amsmath}
\usepackage{graphicx}
\usepackage{diagbox}
\usepackage{rotating}
\usepackage{color}
\usepackage{xspace}
\usepackage{arydshln} 
\usepackage{bbm}

\definecolor{sgadecikir}{rgb}{0.2,0.2,0.22}

\AtBeginDocument{%
  \providecommand\BibTeX{{%
    \normalfont B\kern-0.5em{\scshape i\kern-0.25em b}\kern-0.8em\TeX}}}

\acmConference[CIKM '24]{Proceedings of the 33rd ACM International Conference on Information and Knowledge Management}{October 21--25, 2024}{Boise, ID, USA}
\copyrightyear{2024}
\acmYear{2024}
\setcopyright{acmlicensed}\acmConference[CIKM '24]{Proceedings of the 33rd ACM International Conference on Information and Knowledge Management}{October 21--25, 2024}{Boise, ID, USA}
\acmBooktitle{Proceedings of the 33rd ACM International Conference on Information and Knowledge Management (CIKM '24), October 21--25, 2024, Boise, ID, USA}
\acmDOI{10.1145/3627673.3679934}
\acmISBN{979-8-4007-0436-9/24/10}

\begin{document}



\title{GeoReasoner: Reasoning On Geospatially Grounded Context For Natural Language Understanding}


\author{Yibo Yan}
\affiliation{%
  \institution{National University of Singapore}
  \city{}
  \country{}}
\email{yanyibo70@gmail.com}

\author{Joey Lee}
\affiliation{%
  \institution{Shanghai Zhangjiang Institute of Mathematics}
  \city{}
  \country{}}
\email{joeyleesh82@gmail.com}

\renewcommand{\shortauthors}{Yibo Yan et al.}


\begin{abstract}
In human reading and communication, individuals tend to engage in geospatial reasoning, which involves recognizing geographic entities and making informed inferences about their interrelationships. To mimic such cognitive process, current methods either utilize conventional natural language understanding toolkits, or directly apply models pretrained on geo-related natural language corpora. However, these methods face two significant challenges: i) they do not generalize well to unseen geospatial scenarios, and ii) they overlook the importance of integrating geospatial context from geographical databases with linguistic information from the Internet. To handle these challenges, we propose GeoReasoner, a language model capable of reasoning on geospatially grounded natural language. Specifically, it first leverages Large Language Models (LLMs) to generate a comprehensive location description based on linguistic and geospatial information. It also encodes direction and distance information into spatial embedding via treating them as pseudo-sentences. Consequently, the model is trained on both anchor-level and neighbor-level inputs to learn geo-entity representation. Extensive experimental results demonstrate GeoReasoner's superiority in three tasks: toponym recognition, toponym linking, and geo-entity typing, compared to the state-of-the-art baselines.

\end{abstract}




\maketitle

\section{INTRODUCTION}
\begin{figure}[!t]
  \centering
  \includegraphics[width=\linewidth]{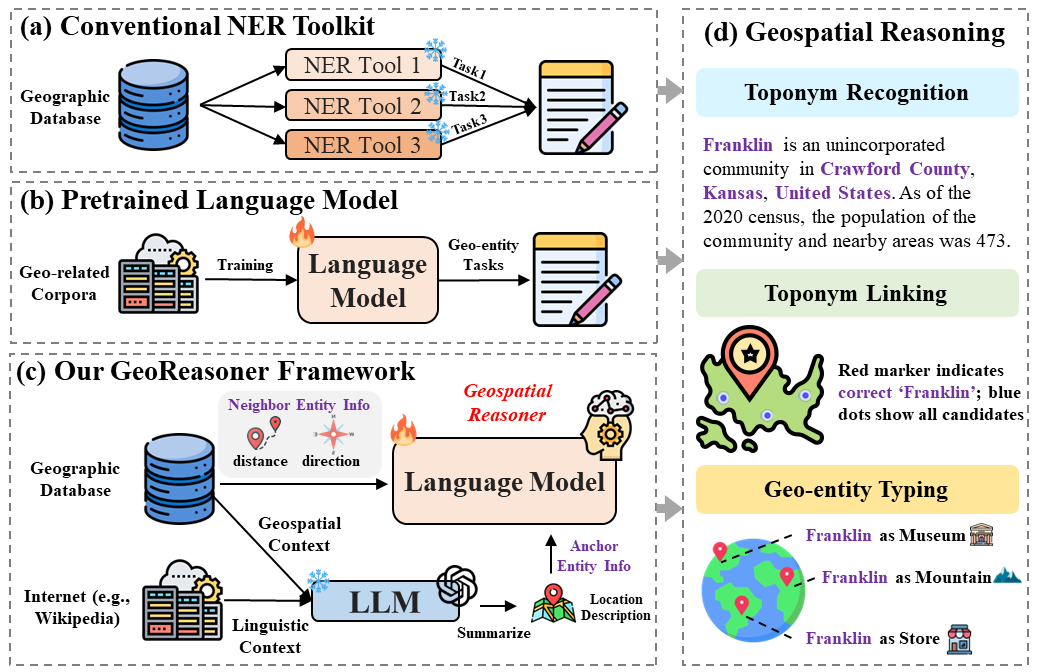}
  \vspace{-2em}
  \caption{Comparison between our LLM-assisted GeoReasoner framework and conventional paradigms for geospatial natural language understanding task.}
  \label{fig: intro}
  \vspace{-1em}
\end{figure}

Geospatial reasoning is crucial for understanding and interpreting natural language texts within the context of spatial relationships \cite{stock2016geospatial,mai2023opportunities}. For example, in the sentence "I traveled to San Jose in California to see the Tech Museum," a human can effortlessly recognize the place names "San Jose" and "California". Additionally, a human can infer that the "Tech Museum" is associated with the technology industry prevalent in the region. In practice, geospatially grounded language understanding involves essential tasks including identifying the geospatial concepts mentioned and deducing the identities of these concepts \cite{li2023geolm,deng2024geoscience,masis2024earth}, as shown in Figure \ref{fig: intro} (d). Therefore, such tasks heavily rely on the one-to-one correspondence between geographic entities and the physical world, especially in real-world location-based services and navigation tasks \cite{hu2022location,gritta2018s,wang2020neurotpr,wallgrun2018geocorpora,yan2023urban}.

The rapid development of natural language understanding (NLU) has significantly impacted various fields, including geospatial reasoning \cite{tucker2024systematic}. NLU advancements have enabled more sophisticated approaches to understanding and interpreting natural language texts within spatial contexts, which have led to two key paradigms for tackling \textit{geospatial reasoning tasks}, as shown in Figure \ref{fig: intro}.

a) \textbf{Conventional NLU toolkits} (e.g., Named Entity Resolution tools) are designed to identify and classify named entities in text, including geographic entities. While these tools are effective, they often struggle to generalize to unseen geospatial scenarios, which limits their applicability in dynamic and diverse contexts \cite{naveen2024geonlu}.

b) \textbf{Pretrained language models on geo-related corpora} are leveraged to understand and generate text based on extensive training data, which includes geographic information \cite{lieberman2010geotagging,mani2010spatialml}. However, despite their advanced capabilities, these models often overlook the importance of combining geospatial context from geographical databases with the rich linguistic information available on the Internet. This lack of integration can result in models that are less effective at reasoning about the spatial relationships and interdependencies between geographic entities \cite{li2023geolm}.

To address the aforementioned challenges, we propose GeoReasoner, a comprehensive framework designed to improve geospatial reasoning by integrating both linguistic and geospatial contexts, as illustrated in Figure \ref{fig: intro} (c). First, Our approach utilizes Large Language Models (LLMs) to generate detailed location descriptions by combining information from geographic databases (e.g., OpenStreetMap) and the Internet (e.g., Wikipedia). Then, GeoReasoner encodes spatial information, such as direction and distance, into spatial embeddings by treating them as pseudo-sentences. This innovative method enables the model to learn robust geo-entity representations through a dual training process that incorporates both geospatial and linguistic data. Eventually, by employing geospatial contrastive loss \cite{oord2018representation} and masked language modeling loss \cite{devlin2018bert}, GeoReasoner achieves superior performance in key geospatial reasoning tasks, including toponym recognition, toponym linking, and geo-entity typing, outperforming current state-of-the-art baselines. In summary, GeoReasoner represents a significant step forward by providing a robust solution for accurately understanding and reasoning on geospatial context in natural language texts.

\vspace{-0.5em}
\section{METHODOLOGY}

\begin{figure*}[t]
  \begin{center}
	\includegraphics[width=0.99\textwidth]{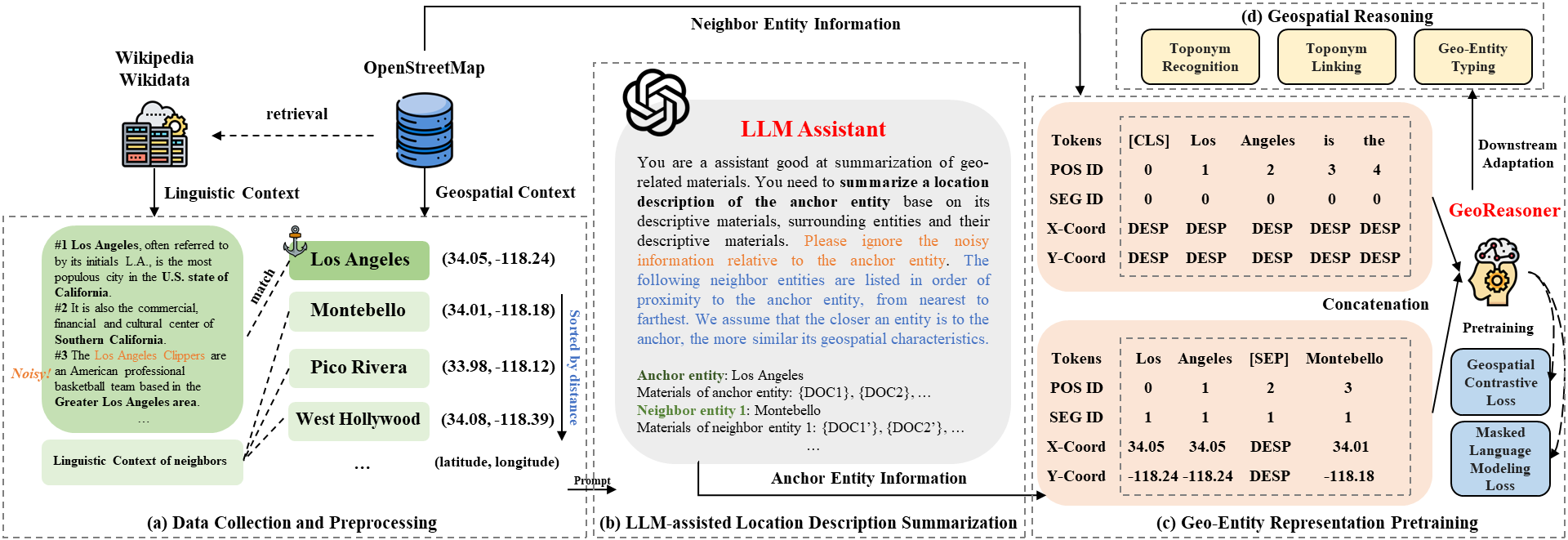}
  \end{center}
  \caption{Overall Framework of GeoReasoner. 
  }\label{fig: framework}
  \vspace{-0.5em}
\end{figure*}

\subsection{Data Collection and Preprocessing}\label{sec: dataset}
We preprocess the training corpora by dividing them into two parts: pseudo-sentence corpora from OpenStreetMap (OSM) to provide geospatial context and natural language corpora from Wikipedia and Wikidata to provide linguistic context. As illustrated in Figure \ref{fig: framework} (a), for the geographical dataset, we use OSM, a crowd-sourced database with extensive point geo-entities worldwide, which links to Wikipedia and Wikidata. We preprocess OSM data to gather geo-entities with these links, creating paired training data for contrastive pretraining. We linearize geospatial context by constructing pseudo-sentences for each geo-entity, sorting neighboring geo-entity names by distance. For the natural language text corpora, we use Wikipedia and Wikidata. We scrape Wikipedia articles linked from OSM annotations, break them into sentences, and use Trie-based phrase matching \cite{hsu2013space} to find sentences containing the corresponding OSM geo-entity names. Training samples are paragraphs with at least one geo-entity name. We collect Wikidata geo-entities via QID identifier and convert relation triples to natural sentences.

\vspace{-0.5em}
\subsection{LLM-assisted Location Description Summarization}\label{sec: summary}
Due to the noisy nature of the linguistic information associated with a given entity (i.e., anchor entity), it is crucial to rewrite it by retaining the most significant parts while integrating the surrounding context. In this phase as shown in Figure \ref{fig: framework} (b), a LLM (specifically GPT-4 Turbo in this study) is employed as a summarizer, utilizing its comprehensive capability to extract key information from both geospatial and linguistic contexts. The geospatial context includes locations and implicit geo-relations of neighboring entities to the anchor geo-entity. Meanwhile, the linguistic context encompasses essential information that specifies a geo-entity, such as sentences describing its geography, environment, culture, and history. Consequently, we define the output of the LLM as a comprehensive location description of the given anchor entity.

\vspace{-0.5em}
\subsection{Geo-Entity Representation Pretraining}\label{sec: learning}
As illustrated in Figure \ref{fig: framework} (c), the training process of GeoReasoner aims to learn and align anchor and neighbor entity information within the same embedding space, thereby obtaining geospatially grounded language representations. Specifically, GeoReasoner uses location descriptions as natural language inputs and generates anchor-level representations by averaging the token representations within the geo-entity name span. For neighbor entity information, GeoReasoner employs the geospatial context linearization method and spatial embedding module from SpaBERT \cite{li2022spabert}, a pretrained language model that contextualizes entity representations using point geographic data. Given a anchor geo-entity and its spatial neighbors, GeoReasoner linearizes the geospatial context by sorting neighbors based on their geospatial distances from the anchor geo-entity. It then concatenates the anchor geo-entity's name with the sorted neighbors to form a pseudo-sentence. To preserve directional relations and relative distances, GeoReasoner uses a geocoordinates embedding module, via a sinusoidal position embedding layer.

To enable GeoReasoner to process both natural language text and geographical data, we leverage a specific position embedding mechanism for each token alongside the token embeddings \cite{li2023geolm}. As shown in Figure \ref{fig: framework} (c), \textit{Position ID} describes the token's index position in the sentence (it starts from zero for both anchor-level and neighbor-level inputs). \textit{Segment ID} indicates the source of the input tokens (tokens from the anchor-level input have a Segment ID of zero, while those from the neighbor-level input have a Segment ID of one). \textit{X-coord and Y-coord} refer to inputs for the spatial coordinate embedding (tokens within the same geo-entity name span share the same X-coord and Y-coord values, but anchor-level tokens, lacking geocoordinate information, have their X-coord and Y-coord set to \texttt{DSEP}, a constant distance filler value). 

During pretraining, two tasks are employed to learn geospatially grounded representations of natural language text. The first task is \textit{geospatial contrastive learning} using an InfoNCE loss \cite{oord2018representation}, which contrasts geo-entity features extracted from different contextual levels. This loss encourages GeoReasoner to generate similar representations for the same geo-entity. Consequently, each batch comprises 50\% random negatives and 50\% hard negatives.

Formally, Define the training data $\mathcal{D}$ consist of pairs of samples $(s^{loc}_i,s^{geo}_i)$, where $s^{loc}_i$ and $s^{geo}_i$ refer to anchor-level location description and neighbor-level pseudo-sentence created from the geospatial context. Both refer to the same geo-entity.
Define $f(\cdot)$ be GeoReasoner that takes both $s^{loc}_i$ and $s^{geo}_i$ as input and generates final representation $\mathbf{h}_i^{loc} = f(s^{loc}_i)$ and $\mathbf{h}_i^{geo} = f(s^{geo}_i)$. Hence, the geospatial contrastive loss $\mathcal{L}_{i}^{con}$ can be defined as:

\begin{equation*}
    \mathcal{L}_{i}^{con} = - \mathrm{log}  
    \frac{e^{\mathrm{sim}(\mathbf{h}_i^{loc}, \mathbf{h}_i^{geo})/\tau}}             {\sum_{j=1}^{2N}\mathbbm{1}_{[j\neq i]}  e^{\mathrm{sim}(\mathbf{h}_i^{loc}, \mathbf{h}_j^{geo})/\tau}} ,
\end{equation*}

\noindent
where $\tau$ and $\mathrm{sim}(\cdot)$ refer to temperature and cosine similarity. 

Additionally, we employ a \textit{masked language modeling} task \cite{devlin2018bert} on a concatenation of the paired anchor-level sentence and neighbor-level pseudo-sentence. This task encourages GeoReasoner to recover masked tokens, thereby improving its ability to integrate and utilize information from both contexts.

\vspace{-0.5em}
\subsection{Geospatial Downstream Tasks}\label{sec: downstream}

Our study further adapts GeoReasoner to the following downstream tasks to demonstrate its ability for geospatially grounded language understanding, as shown in Figure \ref{fig: framework} (d).

\noindent\textbf{Toponym recognition} involves identifying and extracting place names from unstructured text \cite{gritta2020pragmatic}. This is achieved by adding a fully connected layer to GeoReasoner and training the model on a downstream dataset, enabling it to classify each token accurately.

\noindent\textbf{Toponym linking} aims to identify the specific geo-entity mentioned in text by matching it to the correct record in a geographic database \cite{gritta2020pragmatic}. This database may include multiple entities with the same name as the extracted toponym from the text. 

\noindent\textbf{Geo-entity typing} focuses on categorizing location types within a geographical database. We approach this task as a classification problem by attaching a one-layer classification head to GeoReasoner. During training, GeoReasoner predicts the type of a central geo-entity from a subregion. 

\section{EXPERIMENT}

\begin{table*}[t]
\label{tab: task1}
\centering
    {
    \footnotesize
    \setlength{\tabcolsep}{7pt}
        \begin{tabular}{c|ccc:ccc|ccc} 
        \hline \multirow{2}{*}{ \textbf{\textit{GeoWebNews}}} & \multicolumn{3}{c}{Token (B-topo)} & \multicolumn{3}{c}{Token (I-topo)} & \multicolumn{3}{c}{Entity} \\ 
         & \textbf{Prec} & \textbf{Recall} & \textbf{F1}  
          & \textbf{Prec} & \textbf{Recall} & \textbf{F1}  & 
         \textbf{Prec} & \textbf{Recall} & \textbf{F1}  \\ \hline
        BERT &90.00 &89.28 &89.64 &78.55 &79.44 &78.99  &77.03 &83.42 & 80.10\\
        SimCSE-BERT  &83.86 & 90.26 &86.95 &74.61 &82.07 &78.16  &72.76 &83.68  &77.84\\
        SpanBERT &85.98 &88.37 &87.16 &\textbf{86.13} &\textbf{89.19} & \textbf{87.63}  &75.32 &81.16 &78.13\\
        SapBERT &83.12 &88.32 &85.64 &76.26 &81.11 &78.61 &72.48 &80.16 &76.12\\
        GeoLM &\underline{91.15} &\underline{90.43} &\underline{90.79} &79.16 &84.27 &81.63 &\underline{82.18} &\underline{85.67} &\underline{83.89}\\
        \cellcolor{orange!5}  GeoReasoner & \cellcolor{orange!5}\textbf{93.22}  & \cellcolor{orange!5}\textbf{91.27}  & \cellcolor{orange!5}\textbf{92.23} & \cellcolor{orange!5}\underline{81.56}  & \cellcolor{orange!5}\underline{87.90}  & \cellcolor{orange!5}\underline{84.61}  & \cellcolor{orange!5}\textbf{84.66}  & \cellcolor{orange!5}\textbf{85.78} & \cellcolor{orange!5}\textbf{85.17} \\
        
        \hline
        \end{tabular}
    }
\caption{\label{tab:toponym_recognition_results} Result of toponym recognition task on GeoWebNews dataset. \textbf{Bolded} and \underline{underlined} numbers are best and second best scores, respectively. We frame this task as a multi-class sequence tagging problem, where tokens are classified into one of three classes: \textit{B-topo} (beginning of a toponym), \textit{I-topo} (inside a toponym), or \textit{O} (non-toponym).}
\end{table*}

\begin{table*}
\centering
\small
\setlength{\tabcolsep}{7pt}
\begin{tabular}{lcccccccccc}
\hline 
\textbf{Classes } & \textbf{Edu.} & \textbf{Ent.}& \textbf{Fac.}& \textbf{Fin.}& \textbf{Hea.}& \textbf{Pub.}& \textbf{Sus.}& \textbf{Tra.}& \textbf{Was.} & \textbf{Micro F1} \\ 
\hline
BERT  &  67.4 &   63.4 &   \textbf{76.3} &   92.9 &   85.6 &   87.2 &   85.6 &   86.2 &   67.8 & 83.5  \\
SpanBERT    &   63.3 &   58.9 &   60.8 &   91.6 &   85.9 &   \underline{88.2}  &   82.4 &   86.7  &   \textbf{73.5} & 81.9 \\
SimCSE-BERT & 62.3 &   59.0 &   50.4 &   92.5 &   86.7 &   85.2 &   85.7 &   81.0 &   47.0 & 81.0 \\
LUKE  &  64.8 &   60.8 &   59.8 &   94.5 &   85.7 &   86.7 &   85.4 &   85.1 &   51.7 & 82.5 \\
SpaBERT  &    67.4 &    65.3 &    68.0 &     95.9 &     86.5 &     \textbf{90.0} &     88.3 &     88.8 &     \underline{70.3} &   85.2 \\
GeoLM  &  \underline{72.5} &   \underline{70.9} &   73.0 &   \underline{97.8} &   \underline{91.5} &   83.6 &   \underline{90.5} &   \textbf{90.8} &   62.2 & \underline{87.8} \\

\cellcolor{orange!5}GeoReasoner  &  \cellcolor{orange!5}\textbf{73.8} &   \cellcolor{orange!5}\textbf{72.2} &  \cellcolor{orange!5}\underline{76.0} &   \cellcolor{orange!5}\textbf{ 98.0} &   \cellcolor{orange!5}\textbf{93.3} &   \cellcolor{orange!5}87.9 &   \cellcolor{orange!5}\textbf{92.1} &   \cellcolor{orange!5}\underline{90.6} &   \cellcolor{orange!5}65.8 & \cellcolor{orange!5}\textbf{88.9} \\

\hline 

\end{tabular}
\caption{\label{tab:typing_result} Result of geo-entity typing task. Column names are the OSM classes (education, entertainment, facility, financial, healthcare, public service, sustenance, transportation and waste management). \textbf{Bolded} and \underline{underlined} numbers are best and second best scores.}
\end{table*}

\begin{table}[t]
\centering
\footnotesize
\setlength{\tabcolsep}{5pt}
    \begin{tabular}{c|ccc}

    \hline 
     \textbf{\textit{LGL}} & \textbf{R@1} & \textbf{R@5} & \textbf{R@10} \\ \hline
    BERT & 34.6  & \underline{67.5} & \textbf{78.1}\\
    RoBERTa &24.2  & 48.7 & 60.6 \\
    SpanBERT & 25.2 &  48.8 & 61.0 \\
    SapBERT & 30.8  & 58.8  & 72.2\\
    GeoLM & \underline{38.2}  & 65.3  & 72.6 \\
     \cellcolor{orange!5}GeoReasoner & \cellcolor{orange!5}\textbf{40.1} &  \cellcolor{orange!5}\textbf{68.3} & \cellcolor{orange!5}\underline{75.9}\\ 

    \hline 

    \end{tabular}
    
\caption{\label{tab:linking_result} Result of toponym linking task on LGL dataset. \textbf{Bolded} and \underline{underlined} numbers are best and second best scores, respectively. \textbf{R}@$k$ measures whether ground-truth GeoNames ID presents among the top $k$ retrieval results.}
\end{table}

\begin{table}
\centering
\small
\setlength{\tabcolsep}{3pt}
\begin{tabular}{l|ccc}
\hline 
\textbf{Ablation Settings} & \textbf{Prec} & \textbf{Recall}& \textbf{F1} \\ 
\hline

\cellcolor{orange!5}GeoReasoner & \cellcolor{orange!5}84.66& \cellcolor{orange!5}85.78& \cellcolor{orange!5}85.17 \\
GeoReasoner w/o contrastive loss & 73.67 & 79.45 & 76.45\\
GeoReasoner w/o MLM loss & 78.23 & 80.22 & 79.21\\
GeoReasoner w/o spatial embed. & 76.98& 85.31 & 80.93\\
GeoReasoner w/o LLM summarization & 79.12& 83.42& 81.17\\

\hline 

\end{tabular}
\caption{\label{tab:ablation_result} Ablation study on toponym recognition task.}
\vspace{-1em}
\end{table}

\subsection{Experimental Setup}
\subsubsection{Datasets}

For toponym recognition, we utilize the GeoWebNews dataset \cite{gritta2020pragmatic}, which includes 200 news articles containing 2,601 toponyms with valid geocoordinates. For toponym linking, we use the Local Global Corpus \cite{lieberman2010geotagging} with 588 news articles. For geo-entity typing, we employ the dataset from \cite{li2022spabert} with geospatial context for the anchor geo-entity. The train/test ratio is 8:2.

\subsubsection{Evaluation}
For toponym recognition, we report precision, recall, and F1 scores at both token-level and entity-level, with entity-level accuracy requiring exact matches to the ground-truth. In toponym linking, we evaluate the accuracy of linking toponyms to their correct geo-entities in the GeoNames database, emphasizing disambiguation. For geo-entity typing, we follow \cite{li2022spabert} and report F1 score for each amenity class and overall micro F1 score.

\subsubsection{Baselines}
In toponym recognition, we compare GeoReasoner with fine-tuned models such as BERT \cite{devlin2018bert}, SimCSE-BERT \cite{gao2021simcse}, SpanBERT \cite{joshi2020spanbert}, SapBERT \cite{liu2020self} and GeoLM \cite{li2023geolm}. For toponym linking, we benchmark against the same models. In geo-entity typing, we extend comparisons to include LUKE \cite{yamada2020luke} and SpaBERT \cite{li2022spabert}, which leverage specialized tokenization and geospatial context, respectively. All models are evaluated in their \verb|base| versions.

\vspace{-0.5em}
\subsection{Overall Performance}
\subsubsection{Toponym Recognition}
As indicated in Table \ref{tab:toponym_recognition_results}, GeoReasoner shows a balanced and strong performance across both token-level and entity-level metrics, especially excelling in the identification of B-topo tokens. SpanBERT excels in predicting I-topo tokens, which indicates the benefit of span prediction during pretraining for recognizing the continuation of toponyms. The results suggest that models such as GeoReasoner and GeoLM, which are specifically trained with geospatial grounding, outperform general models like BERT and its variants in toponym recognition tasks.

\subsubsection{Geo-Entity Typing}
The results from Table \ref{tab:typing_result} clearly highlight the superior performance of the GeoReasoner and GeoLM models due to their robust performance among most types. This reveals that contrastive learning applied during pretraining has effectively facilitated the alignment of linguistic and geospatial contexts, though only geospatial context is given during inference. GeoReasoner performs worse on waste management class compared to SpanBERT, but it still exceed the performance of GeoLM, which also integrates geospatial context with linguistic context.


\subsubsection{Toponym Linking}
As shown in Table \ref{tab:linking_result}, GeoReasoner indicates the best capability in terms of top-1 retrieval (\textbf{R@1}) and top-5 retrieval (\textbf{R@5}), and the second best performance on top-10 retrieval (\textbf{R@10}). This underscores the advantage of integrating both geospatial and linguistic context in GeoReasoner compared to baseline models that emphasize one over the other. Moreover, since this task specifically necessitates models to rely solely on linguistic context, GeoReasoner's geospatial contrastive learning paradigm proves beneficial by facilitating alignment among sources.

\vspace{-0.6em}
\subsection{Ablation Study} 


Four ablation experiments are conducted on toponym recognition task: i) removing the geospatial contrastive loss; ii) removing the masked language modeling loss; iii) removing the spatial coordinate embedding layer that leverages geo coordinates as input; iv) removing the inclusion of LLM for location description summarization (i.e., linguistic context will be directly used as anchor-level input for GeoReasoner). Results from Table \ref{tab:ablation_result} demonstrate a significant drop in performance when either the contrastive loss or the masked language modeling loss is removed. This underscores their crucial roles during pretraining. Furthermore, all metrics decrease when the spatial coordinate embedding layer is omitted, indicating that geocoordinates encapsulate implicit geospatial context. Additionally, performance declines without LLM-based summarization, likely due to the noisy nature of the original linguistic context.

\section{Related Work}
Research on understanding geospatial concepts in natural language has used general-purpose NER tools like Stanford NER \cite{finkel2005incorporating} and NeuroNER \cite{dernoncourt2017neuroner}, as well as geospatial-specific tools such as the Edinburgh Geoparser \cite{grover2010use, 10.1145/1722080.1722089} and Mordecai \cite{halterman2017mordecai, halterman2023mordecai}, to identify and link toponyms to geographical databases. Deep learning models, including TopoCluster \cite{delozier2015gazetteer} and CamCoder \cite{gritta2018melbourne}, have connected toponyms with geographic locations using geodesic and lexical features. However, these models often lack effective generalizability of unseen geospatial context during inference. Furthermore, SpaBERT \cite{li2022spabert} serves as a language model trained on geographical corpora, but it cannot leverage linguistic information from the Internet to obtain comprehensive geo-entity representation. GeoLM \cite{li2023geolm} manages to fuse geospatial and linguistic information together via projecting them into a joint embedding. Our GeoReasoner framework stands out by harnessing LLMs' powerful summarization abilities to derive anchor entity information as linguistic context, while seamlessly incorporating geospatial context through concatenation.

\section{CONCLUSION}
GeoReasoner effectively addresses the challenges of geospatial reasoning in natural language processing by integrating linguistic information with geospatial context. It leverages powerful LLMs to generate location descriptions and encodes spatial information as pseudo-sentences. In future research, we will focus on integrating geospatial reasoning with the reasoning capabilities of LLMs.

\bibliographystyle{ACM-Reference-Format}
\bibliography{georeasoner_reference}


\begin{thebibliography}{31}


\ifx \showCODEN    \undefined \def \showCODEN     #1{\unskip}     \fi
\ifx \showDOI      \undefined \def \showDOI       #1{#1}\fi
\ifx \showISBNx    \undefined \def \showISBNx     #1{\unskip}     \fi
\ifx \showISBNxiii \undefined \def \showISBNxiii  #1{\unskip}     \fi
\ifx \showISSN     \undefined \def \showISSN      #1{\unskip}     \fi
\ifx \showLCCN     \undefined \def \showLCCN      #1{\unskip}     \fi
\ifx \shownote     \undefined \def \shownote      #1{#1}          \fi
\ifx \showarticletitle \undefined \def \showarticletitle #1{#1}   \fi
\ifx \showURL      \undefined \def \showURL       {\relax}        \fi
\providecommand\bibfield[2]{#2}
\providecommand\bibinfo[2]{#2}
\providecommand\natexlab[1]{#1}
\providecommand\showeprint[2][]{arXiv:#2}

\bibitem[DeLozier et~al\mbox{.}(2015)]%
        {delozier2015gazetteer}
\bibfield{author}{\bibinfo{person}{Grant DeLozier}, \bibinfo{person}{Jason Baldridge}, {and} \bibinfo{person}{Loretta London}.} \bibinfo{year}{2015}\natexlab{}.
\newblock \showarticletitle{Gazetteer-independent toponym resolution using geographic word profiles}. In \bibinfo{booktitle}{\emph{Proceedings of the AAAI Conference on Artificial Intelligence}}, Vol.~\bibinfo{volume}{29}.
\newblock


\bibitem[Deng et~al\mbox{.}(2024)]%
        {deng2024geoscience}
\bibfield{author}{\bibinfo{person}{Cheng Deng}, \bibinfo{person}{Le Zhou}, \bibinfo{person}{Yi Xu}, \bibinfo{person}{Tianhang Zhang}, \bibinfo{person}{Zhouhan Lin}, \bibinfo{person}{Xinbing Wang}, {and} \bibinfo{person}{Chenghu Zhou}.} \bibinfo{year}{2024}\natexlab{}.
\newblock \bibinfo{booktitle}{\emph{Geoscience Knowledge Understanding and Utilization via Data-centric Large Language Model}}.
\newblock \bibinfo{type}{{T}echnical {R}eport}. \bibinfo{institution}{Copernicus Meetings}.
\newblock


\bibitem[Dernoncourt et~al\mbox{.}(2017)]%
        {dernoncourt2017neuroner}
\bibfield{author}{\bibinfo{person}{Franck Dernoncourt}, \bibinfo{person}{Ji~Young Lee}, {and} \bibinfo{person}{Peter Szolovits}.} \bibinfo{year}{2017}\natexlab{}.
\newblock \showarticletitle{NeuroNER: an easy-to-use program for named-entity recognition based on neural networks}.
\newblock \bibinfo{journal}{\emph{arXiv preprint arXiv:1705.05487}} (\bibinfo{year}{2017}).
\newblock


\bibitem[Devlin et~al\mbox{.}(2018)]%
        {devlin2018bert}
\bibfield{author}{\bibinfo{person}{Jacob Devlin}, \bibinfo{person}{Ming-Wei Chang}, \bibinfo{person}{Kenton Lee}, {and} \bibinfo{person}{Kristina Toutanova}.} \bibinfo{year}{2018}\natexlab{}.
\newblock \showarticletitle{Bert: Pre-training of deep bidirectional transformers for language understanding}.
\newblock \bibinfo{journal}{\emph{arXiv preprint arXiv:1810.04805}} (\bibinfo{year}{2018}).
\newblock


\bibitem[Finkel et~al\mbox{.}(2005)]%
        {finkel2005incorporating}
\bibfield{author}{\bibinfo{person}{Jenny~Rose Finkel}, \bibinfo{person}{Trond Grenager}, {and} \bibinfo{person}{Christopher~D Manning}.} \bibinfo{year}{2005}\natexlab{}.
\newblock \showarticletitle{Incorporating non-local information into information extraction systems by gibbs sampling}. In \bibinfo{booktitle}{\emph{Proceedings of the 43rd annual meeting of the association for computational linguistics (ACL’05)}}. \bibinfo{pages}{363--370}.
\newblock


\bibitem[Gao et~al\mbox{.}(2021)]%
        {gao2021simcse}
\bibfield{author}{\bibinfo{person}{Tianyu Gao}, \bibinfo{person}{Xingcheng Yao}, {and} \bibinfo{person}{Danqi Chen}.} \bibinfo{year}{2021}\natexlab{}.
\newblock \showarticletitle{Simcse: Simple contrastive learning of sentence embeddings}.
\newblock \bibinfo{journal}{\emph{arXiv preprint arXiv:2104.08821}} (\bibinfo{year}{2021}).
\newblock


\bibitem[Gritta et~al\mbox{.}(2018a)]%
        {gritta2018melbourne}
\bibfield{author}{\bibinfo{person}{Milan Gritta}, \bibinfo{person}{Mohammad~Taher Pilehvar}, {and} \bibinfo{person}{Nigel Collier}.} \bibinfo{year}{2018}\natexlab{a}.
\newblock \showarticletitle{Which melbourne? augmenting geocoding with maps}. Association for Computational Linguistics.
\newblock


\bibitem[Gritta et~al\mbox{.}(2020)]%
        {gritta2020pragmatic}
\bibfield{author}{\bibinfo{person}{Milan Gritta}, \bibinfo{person}{Mohammad~Taher Pilehvar}, {and} \bibinfo{person}{Nigel Collier}.} \bibinfo{year}{2020}\natexlab{}.
\newblock \showarticletitle{A pragmatic guide to geoparsing evaluation: Toponyms, Named Entity Recognition and pragmatics}.
\newblock \bibinfo{journal}{\emph{Language resources and evaluation}}  \bibinfo{volume}{54} (\bibinfo{year}{2020}), \bibinfo{pages}{683--712}.
\newblock


\bibitem[Gritta et~al\mbox{.}(2018b)]%
        {gritta2018s}
\bibfield{author}{\bibinfo{person}{Milan Gritta}, \bibinfo{person}{Mohammad~Taher Pilehvar}, \bibinfo{person}{Nut Limsopatham}, {and} \bibinfo{person}{Nigel Collier}.} \bibinfo{year}{2018}\natexlab{b}.
\newblock \showarticletitle{What’s missing in geographical parsing?}
\newblock \bibinfo{journal}{\emph{Language Resources and Evaluation}}  \bibinfo{volume}{52} (\bibinfo{year}{2018}), \bibinfo{pages}{603--623}.
\newblock


\bibitem[Grover et~al\mbox{.}(2010)]%
        {grover2010use}
\bibfield{author}{\bibinfo{person}{Claire Grover}, \bibinfo{person}{Richard Tobin}, \bibinfo{person}{Kate Byrne}, \bibinfo{person}{Matthew Woollard}, \bibinfo{person}{James Reid}, \bibinfo{person}{Stuart Dunn}, {and} \bibinfo{person}{Julian Ball}.} \bibinfo{year}{2010}\natexlab{}.
\newblock \showarticletitle{Use of the Edinburgh geoparser for georeferencing digitized historical collections}.
\newblock \bibinfo{journal}{\emph{Philosophical Transactions of the Royal Society A: Mathematical, Physical and Engineering Sciences}} \bibinfo{volume}{368}, \bibinfo{number}{1925} (\bibinfo{year}{2010}), \bibinfo{pages}{3875--3889}.
\newblock


\bibitem[Halterman(2017)]%
        {halterman2017mordecai}
\bibfield{author}{\bibinfo{person}{Andrew Halterman}.} \bibinfo{year}{2017}\natexlab{}.
\newblock \showarticletitle{Mordecai: Full Text Geoparsing and Event Geocoding}.
\newblock \bibinfo{journal}{\emph{The Journal of Open Source Software}} \bibinfo{volume}{2}, \bibinfo{number}{9} (\bibinfo{year}{2017}).
\newblock
\urldef\tempurl%
\url{https://doi.org/10.21105/joss.00091}
\showDOI{\tempurl}


\bibitem[Halterman(2023)]%
        {halterman2023mordecai}
\bibfield{author}{\bibinfo{person}{Andrew Halterman}.} \bibinfo{year}{2023}\natexlab{}.
\newblock \showarticletitle{Mordecai 3: A Neural Geoparser and Event Geocoder}.
\newblock \bibinfo{journal}{\emph{arXiv preprint arXiv:2303.13675}} (\bibinfo{year}{2023}).
\newblock


\bibitem[Hsu and Ottaviano(2013)]%
        {hsu2013space}
\bibfield{author}{\bibinfo{person}{Bo-June Hsu} {and} \bibinfo{person}{Giuseppe Ottaviano}.} \bibinfo{year}{2013}\natexlab{}.
\newblock \showarticletitle{Space-efficient data structures for top-k completion}. In \bibinfo{booktitle}{\emph{Proceedings of the 22nd international conference on World Wide Web}}. \bibinfo{pages}{583--594}.
\newblock


\bibitem[Hu et~al\mbox{.}(2022)]%
        {hu2022location}
\bibfield{author}{\bibinfo{person}{Xuke Hu}, \bibinfo{person}{Zhiyong Zhou}, \bibinfo{person}{Hao Li}, \bibinfo{person}{Yingjie Hu}, \bibinfo{person}{Fuqiang Gu}, \bibinfo{person}{Jens Kersten}, \bibinfo{person}{Hongchao Fan}, {and} \bibinfo{person}{Friederike Klan}.} \bibinfo{year}{2022}\natexlab{}.
\newblock \showarticletitle{Location reference recognition from texts: A survey and comparison}.
\newblock \bibinfo{journal}{\emph{arXiv preprint arXiv:2207.01683}} (\bibinfo{year}{2022}).
\newblock


\bibitem[Joshi et~al\mbox{.}(2020)]%
        {joshi2020spanbert}
\bibfield{author}{\bibinfo{person}{Mandar Joshi}, \bibinfo{person}{Danqi Chen}, \bibinfo{person}{Yinhan Liu}, \bibinfo{person}{Daniel~S Weld}, \bibinfo{person}{Luke Zettlemoyer}, {and} \bibinfo{person}{Omer Levy}.} \bibinfo{year}{2020}\natexlab{}.
\newblock \showarticletitle{Spanbert: Improving pre-training by representing and predicting spans}.
\newblock \bibinfo{journal}{\emph{Transactions of the association for computational linguistics}}  \bibinfo{volume}{8} (\bibinfo{year}{2020}), \bibinfo{pages}{64--77}.
\newblock


\bibitem[Li et~al\mbox{.}(2022)]%
        {li2022spabert}
\bibfield{author}{\bibinfo{person}{Zekun Li}, \bibinfo{person}{Jina Kim}, \bibinfo{person}{Yao-Yi Chiang}, {and} \bibinfo{person}{Muhao Chen}.} \bibinfo{year}{2022}\natexlab{}.
\newblock \showarticletitle{Spabert: a pretrained language model from geographic data for geo-entity representation}.
\newblock \bibinfo{journal}{\emph{arXiv preprint arXiv:2210.12213}} (\bibinfo{year}{2022}).
\newblock


\bibitem[Li et~al\mbox{.}(2023)]%
        {li2023geolm}
\bibfield{author}{\bibinfo{person}{Zekun Li}, \bibinfo{person}{Wenxuan Zhou}, \bibinfo{person}{Yao-Yi Chiang}, {and} \bibinfo{person}{Muhao Chen}.} \bibinfo{year}{2023}\natexlab{}.
\newblock \showarticletitle{Geolm: Empowering language models for geospatially grounded language understanding}.
\newblock \bibinfo{journal}{\emph{arXiv preprint arXiv:2310.14478}} (\bibinfo{year}{2023}).
\newblock


\bibitem[Lieberman et~al\mbox{.}(2010)]%
        {lieberman2010geotagging}
\bibfield{author}{\bibinfo{person}{Michael~D Lieberman}, \bibinfo{person}{Hanan Samet}, {and} \bibinfo{person}{Jagan Sankaranarayanan}.} \bibinfo{year}{2010}\natexlab{}.
\newblock \showarticletitle{Geotagging with local lexicons to build indexes for textually-specified spatial data}. In \bibinfo{booktitle}{\emph{2010 IEEE 26th international conference on data engineering (ICDE 2010)}}. IEEE, \bibinfo{pages}{201--212}.
\newblock


\bibitem[Liu et~al\mbox{.}(2020)]%
        {liu2020self}
\bibfield{author}{\bibinfo{person}{Fangyu Liu}, \bibinfo{person}{Ehsan Shareghi}, \bibinfo{person}{Zaiqiao Meng}, \bibinfo{person}{Marco Basaldella}, {and} \bibinfo{person}{Nigel Collier}.} \bibinfo{year}{2020}\natexlab{}.
\newblock \showarticletitle{Self-alignment pretraining for biomedical entity representations}.
\newblock \bibinfo{journal}{\emph{arXiv preprint arXiv:2010.11784}} (\bibinfo{year}{2020}).
\newblock


\bibitem[Mai et~al\mbox{.}(2023)]%
        {mai2023opportunities}
\bibfield{author}{\bibinfo{person}{Gengchen Mai}, \bibinfo{person}{Weiming Huang}, \bibinfo{person}{Jin Sun}, \bibinfo{person}{Suhang Song}, \bibinfo{person}{Deepak Mishra}, \bibinfo{person}{Ninghao Liu}, \bibinfo{person}{Song Gao}, \bibinfo{person}{Tianming Liu}, \bibinfo{person}{Gao Cong}, \bibinfo{person}{Yingjie Hu}, {et~al\mbox{.}}} \bibinfo{year}{2023}\natexlab{}.
\newblock \showarticletitle{On the opportunities and challenges of foundation models for geospatial artificial intelligence}.
\newblock \bibinfo{journal}{\emph{arXiv preprint arXiv:2304.06798}} (\bibinfo{year}{2023}).
\newblock


\bibitem[Mani et~al\mbox{.}(2010)]%
        {mani2010spatialml}
\bibfield{author}{\bibinfo{person}{Inderjeet Mani}, \bibinfo{person}{Christy Doran}, \bibinfo{person}{Dave Harris}, \bibinfo{person}{Janet Hitzeman}, \bibinfo{person}{Rob Quimby}, \bibinfo{person}{Justin Richer}, \bibinfo{person}{Ben Wellner}, \bibinfo{person}{Scott Mardis}, {and} \bibinfo{person}{Seamus Clancy}.} \bibinfo{year}{2010}\natexlab{}.
\newblock \showarticletitle{SpatialML: annotation scheme, resources, and evaluation}.
\newblock \bibinfo{journal}{\emph{Language Resources and Evaluation}}  \bibinfo{volume}{44} (\bibinfo{year}{2010}), \bibinfo{pages}{263--280}.
\newblock


\bibitem[Masis and O'Connor(2024)]%
        {masis2024earth}
\bibfield{author}{\bibinfo{person}{Tessa Masis} {and} \bibinfo{person}{Brendan O'Connor}.} \bibinfo{year}{2024}\natexlab{}.
\newblock \showarticletitle{Where on Earth Do Users Say They Are?: Geo-Entity Linking for Noisy Multilingual User Input}.
\newblock \bibinfo{journal}{\emph{arXiv:2404.18784}} (\bibinfo{year}{2024}).
\newblock


\bibitem[Naveen et~al\mbox{.}(2024)]%
        {naveen2024geonlu}
\bibfield{author}{\bibinfo{person}{Palanichamy Naveen}, \bibinfo{person}{Rajagopal Maheswar}, {and} \bibinfo{person}{Pavel Trojovsk{\`y}}.} \bibinfo{year}{2024}\natexlab{}.
\newblock \showarticletitle{GeoNLU: Bridging the gap between natural language and spatial data infrastructures}.
\newblock \bibinfo{journal}{\emph{Alexandria Engineering Journal}}  \bibinfo{volume}{87} (\bibinfo{year}{2024}), \bibinfo{pages}{126--147}.
\newblock


\bibitem[Oord et~al\mbox{.}(2018)]%
        {oord2018representation}
\bibfield{author}{\bibinfo{person}{Aaron van~den Oord}, \bibinfo{person}{Yazhe Li}, {and} \bibinfo{person}{Oriol Vinyals}.} \bibinfo{year}{2018}\natexlab{}.
\newblock \showarticletitle{Representation learning with contrastive predictive coding}.
\newblock \bibinfo{journal}{\emph{arXiv preprint arXiv:1807.03748}} (\bibinfo{year}{2018}).
\newblock


\bibitem[Stock and Guesgen(2016)]%
        {stock2016geospatial}
\bibfield{author}{\bibinfo{person}{Kristin Stock} {and} \bibinfo{person}{Hans Guesgen}.} \bibinfo{year}{2016}\natexlab{}.
\newblock \showarticletitle{Geospatial reasoning with open data}.
\newblock In \bibinfo{booktitle}{\emph{Automating open source intelligence}}. \bibinfo{publisher}{Elsevier}, \bibinfo{pages}{171--204}.
\newblock


\bibitem[Tobin et~al\mbox{.}(2010)]%
        {10.1145/1722080.1722089}
\bibfield{author}{\bibinfo{person}{Richard Tobin}, \bibinfo{person}{Claire Grover}, \bibinfo{person}{Kate Byrne}, \bibinfo{person}{James Reid}, {and} \bibinfo{person}{Jo Walsh}.} \bibinfo{year}{2010}\natexlab{}.
\newblock \showarticletitle{Evaluation of Georeferencing}. In \bibinfo{booktitle}{\emph{Proceedings of the 6th Workshop on Geographic Information Retrieval}} (Zurich, Switzerland) \emph{(\bibinfo{series}{GIR '10})}. \bibinfo{publisher}{Association for Computing Machinery}, \bibinfo{address}{New York, NY, USA}, Article \bibinfo{articleno}{7}, \bibinfo{numpages}{8}~pages.
\newblock
\showISBNx{9781605588261}
\urldef\tempurl%
\url{https://doi.org/10.1145/1722080.1722089}
\showDOI{\tempurl}


\bibitem[Tucker(2024)]%
        {tucker2024systematic}
\bibfield{author}{\bibinfo{person}{Sean Tucker}.} \bibinfo{year}{2024}\natexlab{}.
\newblock \showarticletitle{A systematic review of geospatial location embedding approaches in large language models: A path to spatial AI systems}.
\newblock \bibinfo{journal}{\emph{arXiv preprint arXiv:2401.10279}} (\bibinfo{year}{2024}).
\newblock


\bibitem[Wallgr{\"u}n et~al\mbox{.}(2018)]%
        {wallgrun2018geocorpora}
\bibfield{author}{\bibinfo{person}{Jan~Oliver Wallgr{\"u}n}, \bibinfo{person}{Morteza Karimzadeh}, \bibinfo{person}{Alan~M MacEachren}, {and} \bibinfo{person}{Scott Pezanowski}.} \bibinfo{year}{2018}\natexlab{}.
\newblock \showarticletitle{GeoCorpora: building a corpus to test and train microblog geoparsers}.
\newblock \bibinfo{journal}{\emph{International Journal of Geographical Information Science}} \bibinfo{volume}{32}, \bibinfo{number}{1} (\bibinfo{year}{2018}), \bibinfo{pages}{1--29}.
\newblock


\bibitem[Wang et~al\mbox{.}(2020)]%
        {wang2020neurotpr}
\bibfield{author}{\bibinfo{person}{Jimin Wang}, \bibinfo{person}{Yingjie Hu}, {and} \bibinfo{person}{Kenneth Joseph}.} \bibinfo{year}{2020}\natexlab{}.
\newblock \showarticletitle{NeuroTPR: A neuro-net toponym recognition model for extracting locations from social media messages}.
\newblock \bibinfo{journal}{\emph{Transactions in GIS}} \bibinfo{volume}{24}, \bibinfo{number}{3} (\bibinfo{year}{2020}), \bibinfo{pages}{719--735}.
\newblock


\bibitem[Yamada et~al\mbox{.}(2020)]%
        {yamada2020luke}
\bibfield{author}{\bibinfo{person}{Ikuya Yamada}, \bibinfo{person}{Akari Asai}, \bibinfo{person}{Hiroyuki Shindo}, \bibinfo{person}{Hideaki Takeda}, {and} \bibinfo{person}{Yuji Matsumoto}.} \bibinfo{year}{2020}\natexlab{}.
\newblock \showarticletitle{LUKE: Deep contextualized entity representations with entity-aware self-attention}.
\newblock \bibinfo{journal}{\emph{arXiv preprint arXiv:2010.01057}} (\bibinfo{year}{2020}).
\newblock


\bibitem[Yan et~al\mbox{.}(2024)]%
        {yan2023urban}
\bibfield{author}{\bibinfo{person}{Yibo Yan}, \bibinfo{person}{Haomin Wen}, \bibinfo{person}{Siru Zhong}, \bibinfo{person}{Wei Chen}, \bibinfo{person}{Haodong Chen}, \bibinfo{person}{Qingsong Wen}, \bibinfo{person}{Roger Zimmermann}, {and} \bibinfo{person}{Yuxuan Liang}.} \bibinfo{year}{2024}\natexlab{}.
\newblock \showarticletitle{Urbanclip: Learning text-enhanced urban region profilig with contrastive language-image pretraining from the web}. In \bibinfo{booktitle}{\emph{Proceedings of the ACM on Web Conference 2024}}. \bibinfo{pages}{4006--4017}.
\newblock


\end{thebibliography}

\end{document}